\documentclass[letterpaper]{article}
\usepackage{aaai2027} 
\usepackage[hyphens]{url}  
\usepackage{graphicx} 
\usepackage{natbib} 
\usepackage{caption}
\usepackage{algorithm}
\usepackage{algorithmic}

\usepackage{newfloat}
\usepackage{listings}
\DeclareCaptionStyle{ruled}{labelfont=normalfont,labelsep=colon,strut=off} 
\floatstyle{ruled}
\newfloat{listing}{tb}{lst}{}
\floatname{listing}{Listing}

\usepackage{booktabs}

\usepackage{xcolor,colortbl}
\usepackage[mathscr]{euscript}
 \let\mathscr\relax
\usepackage{amsthm}
\usepackage{amsmath}
\usepackage{amssymb}
\usepackage{calc}
\usepackage{amsfonts}
\usepackage{mathrsfs} 
\usepackage{stmaryrd}
\usepackage{multirow}
\usepackage{caption}
\usepackage{subcaption}
\usepackage{xspace}
\usepackage{dsfont}

\newcommand{\hff}{h^{\mathrm{FF}}}
\newcommand{\name}{I2G\@\xspace}
\newcommand{\mpd}{\operatorname{MPD}_{i,n}}

\lstdefinestyle{pddlconstraint}{%
    basicstyle=\ttfamily\small,
    breaklines=true,
    columns=fixed,
    numbers=none,
    xleftmargin=0pt,
    showstringspaces=false,
    frame=none,
    aboveskip=\abovedisplayskip,
    belowskip=\belowdisplayskip,
    morecomment=[l]{;;},
    commentstyle=\itshape\color{black!55},
    morekeywords={exists,forall,and,or,not,implies,count},
    keywordstyle=\bfseries,
    literate={_initial}{{\raisebox{-0.25ex}{{\fontsize{7.5pt}{9pt}\selectfont I}}}}{1}%
             {_goal}{{\raisebox{-0.25ex}{{\fontsize{7.5pt}{9pt}\selectfont G}}}}{1}%
             {_I}{{\raisebox{-0.25ex}{{\fontsize{7.5pt}{9pt}\selectfont I}}}}{1}%
             {_G}{{\raisebox{-0.25ex}{{\fontsize{7.5pt}{9pt}\selectfont G}}}}{1},%
}

\lstdefinestyle{prompt}{%
    basicstyle=\ttfamily\small,
    breaklines=true,
    columns=fixed,
    numbers=none,
    xleftmargin=0pt,
    showstringspaces=false,
    frame=single,
    aboveskip=\abovedisplayskip,
    belowskip=\belowdisplayskip,
    literate={...}{{$\dots$}}3
}

\title{Generating Instance Generators in PDDL Planning}

\author {
    Nicola J. M\"{u}ller\textsuperscript{\rm 1,2,3},
    Naya Rudolph\textsuperscript{\rm 1,2},
    Katharina Stein\textsuperscript{\rm 2},
    J\"{o}rg Hoffmann\textsuperscript{\rm 1,2},
    Ayal Taitler\textsuperscript{\rm 4},
    Timo P. Gros\textsuperscript{\rm 1,2,3}
}
\affiliations {
    \textsuperscript{\rm 1}German Research Center for Artificial Intelligence (DFKI), Saarbr\"{u}cken, Germany \\
    \textsuperscript{\rm 2}Saarland University, Saarland Informatics Campus, Saarbr\"{u}cken, Germany \\
    \textsuperscript{\rm 3}Center for European Research in Trusted Artificial Intelligence (CERTAIN) \\
    \textsuperscript{\rm 4}Ben-Gurion University of the Negev \\
    \{Nicola.Mueller, Naya.Rudolph, Timo\_Phillip.Gros\}@dfki.de,
    kstein@lst.uni-saarland.de,
    hoffmann@cs.uni-saarland.de,
    ataitler@bgu.ac.il
}

\nocopyright
\begin{document}

\maketitle

\begin{abstract}
PDDL, the de-facto standard language in the AI Planning community, is designed to specify planning domains: sets of instances that share the same predicates and action schemas. Yet it does not provide any means to specify the actual instance set, i.e., legality constraints on initial states and goal conditions, as well as possibly domain subset constraints specifying an instance subset we are interested in. One consequence of this is that instance generation has always been ad-hoc, with manually written domain- and subset-specific instance generators. Recent work has started to address this, through reasoning and learning methods that however suffer from scalability limitations. Here we introduce an alternative approach, leveraging LLMs to generate instance-generation programs, with built-in soundness guarantees through prescribed checks. We show that these automatically generated instance generators return large numbers of sound and diverse instances efficiently.
\end{abstract}


PDDL, the planning domain definition language, is the de-facto standard language in the AI Planning community~\cite{mcdermott20001998, bacchus2000subset, long20033rd, hoffmann2005deterministic, gerevini2009deterministic, haslum2019introduction}.
PDDL is designed to specify planning \emph{domains}: typically infinite sets of instances that share the same predicates and action schemas. 
Yet, PDDL does not provide any means to specify the intended set of instances. 

First, in most domains, not all possible instances define legal initial states or goal conditions.
For example, in the Blocksworld domain, a block must not be located on itself; in transportation domains, a truck must be in exactly one location, etc. 
Second, in many settings, we are actually interested in a subset of the legal domain instances; for example, we may want the goal to be a single stack of all blocks, or we may want to specify goal locations only for packages and not for trucks. 
We refer here to the former as \emph{legality constraints}, and to the latter as \emph{subset constraints}.
This distinction has not generally been made in the literature. It is purely conceptual in that both are constraints on initial states, goals and combinations thereof. The distinction makes practical sense however, as legality constraints are fixed once per-domain, whereas subset constraints may vary depending on purpose.

In the absence of a specification language for such constraints in PDDL, past approaches to instance generation have been mostly ad-hoc.
Researchers have manually written instance generators, domain-specific programs that generate some subset of domain instances, with the underlying legality and subset constraints being implicit in the code. 
This opacity has long been a hindrance to identifying how a domain is actually defined, necessitating statements like "the set of instances that can be generated by generator X" when making claims about a domain (e.g.\ \cite{helmert2003complexity, hoffmann2005ignoring, stein2026improved}). 
The lack of a more general methodology has also entailed that instance generators needed to be painstakingly implemented to cater for new domains or instance subsets. 

An early attempt to address this problem was the work by \citet{haslum-scholz:icaps03ws}, which proposed a language allowing to specify state invariants among others. This language, however, has never been adopted by the community.
Later, \citet{fuentetaja2012planning} introduced the first domain-independent instance generation method, including a language to specify legality constraints, with instance generation formulated as a planning problem. Yet the legality language is limited (each constraint on only a single predicate) and no experiments were published.%
\footnote{More work has been done on automated instance generation within domains, adjusting instance size/hardness automatically as suited, e.g., for learning purposes or for use in planning competitions (e.g.\ \cite{fern2004learning,marom2020utilising,torralba2021automatic}).}

Recently, domain-independent instance generation has been addressed in two new approaches.
\citet{grundke2025domain} specify legality constraints as PDDL axioms and use answer set programming to automatically generate instances.
\citet{nunez2024nesig,nunez2025automated} formulate the problem of generating diverse and difficult instances as a Markov Decision Process, and use deep reinforcement learning (RL) to train policies that generate initial states and goals.
While these works constitute significant progress, a weakness of both these approaches are scalability limitations. 
Answer set programming scales worst-case exponentially in the size of the generated instances, and thus quickly reaches its computational limits in practice. 
RL requires days of training time for each domain subset in practice; and the learned instance-generation policy has to generalize from small training instances to the large instances we desire to generate, which is challenging.

In this work, we introduce an alternative approach that leverages LLMs to generate instance-generation programs in Python. 
These automatically generated Python programs are specific to the domain and desired instance subset, and tend to be efficient. 
The soundness of the generated instances is guaranteed through a prescribed post-processing procedure checking legality and subset constraints. 
Legality constraints are implemented \emph{once-per-domain} by the user in the form of Python tests, supporting efficient checks.
For subset constraints, we support declarative specification in first-order logic, facilitating easier changes and maintenance.
Given these inputs, we prompt an LLM to generate an instance generator as a standalone Python module.
We then go through a debugging loop to improve that module (loosely following ideas proposed for per-domain planner generation \cite{silver2024generalized}).
We test the instance generator for a large number of instances across multiple sizes, verifying whether these satisfy the constraints. 
The results are fed back to the LLM as part of a debug prompt, and the loop repeats.
After several iterations, we return the generator that balances instance soundness with diversity (with the latter being inspired by \citet{nunez2024nesig, nunez2025automated}). 
At deployment time, the post-processing procedure guarantees that only sound instances are returned.

We show experimentally that our approach yields sound and diverse instance generators within minutes. 
Further, we show that the resulting instance generators can generate large numbers of instances much more efficiently than the approaches of \citet{grundke2025domain} and \citet{nunez2024nesig,nunez2025automated}.

Moreover, we conduct a case study of instance generation for the purpose of evaluating generalized planning methods, which try to compute plans that generalize across all legal instances of a domain~\cite{bonet2019learning, staahlberg2022learning, staahlberg2026first, silver2024generalized, stein2026improved}. 
Generalized plans tend to differ in their generalization capacity, in particular in the \emph{subset of instances} they work well on. 
Our methodology permits, for the first time, to evaluate these aspects efficiently without excessive manual coding work or generator-learning. 
Our case study shows that interesting insights can be gained in this manner.


\section{Background}
\label{sec:background}
We provide a brief overview of PDDL planning and related work on domain-independent instance generation.

\paragraph{PDDL Planning.}
The Planning Domain Definition Language (PDDL) is a lifted representation for
planning problems, where problems consist of a domain $D$ and a problem instance $I$~\cite{haslum2019introduction}.
The domain $D$ defines a set of predicates $P \in \mathcal{P}$ and a set of action schemas $A \in \mathcal{A}$ that define arguments, preconditions, effects, and action costs.
The instance $I$ defines a set of objects $o \in \mathcal{O}$, an initial state $Init$, and goal conditions $\mathcal{G}$.
By grounding the predicates and action schemas of $D$ with the objects in $I$, we get a problem $\langle \mathcal{S}, s_{0}, \mathcal{S}_{\mathcal{G}}, Act, f \rangle$, which consists of a set of states $s \in \mathcal{S}$, the initial state $s_{0}$, the set of goal states $\mathcal{S}_{\mathcal{G}}$, the sets of applicable ground actions for each state $Act(s)$, and the transition function $f: \mathcal{S} \times Act \rightarrow \mathcal{S}$~\cite{ghallab2004automated}.
A plan for an instance $I$ is a sequence of ground actions $\vec{a} = \langle a_{0}, \dots, a_{T-1} \rangle$ that transitions from the initial state $s_{0}$ to a goal state $s_{T} \in \mathcal{S}_{\mathcal{G}}$.

\paragraph{Related Work.}
\citet{grundke2025domain} propose a domain-independent instance-generation approach that operates on formal specifications of PDDL domains, which use axioms to characterize legal initial states and goals.
We refer to this approach as ``Gr25''. 
Given a desired number of objects, Gr25 translates the formal domain specification into an answer set program (ASP) whose answer sets correspond to legal planning instances. 
An off-the-shelf ASP solver is then used to compute a set of answer sets, corresponding to legal instances.

The NeSIG approach by \citet{nunez2024nesig} formulates instance generation as a sequential decision-making problem and learns separate reinforcement-learning policies for constructing initial states and goals.
At each step, a policy adds an object or fact to the partially generated instance or terminates the corresponding generation phase. 
During this process, user-provided consistency rules check whether the instance's initial state is legal.
NeSIG's reward function encourages the policies to construct instances that are legal, diverse, and difficult.
To compute the diversity, NeSIG computes a set of features for each constructed instance and then computes the distance between these feature vectors.
To compute the difficulty, NeSIG runs a set of existing planners on the constructed problems and measures the number of expanded search nodes.


\section{Generation of Instance Generators}
We now introduce our approach for generating instance generators in PDDL planning, which we call \name (\textbf{I}nstance \textbf{G}enerator \textbf{G}eneration).
Our approach consists of four steps:
First, we prompt an LLM to generate an instance generator.
Second, we test the resulting instance generator.
Third, based on the test results, we prompt the LLM to improve its generator and go back to step two.
Lastly, after several iterations of improvement, we select the final generator as the one that balances soundness and diversity.

\subsection{LLM Prompting}
To generate an initial instance generator, we construct an LLM prompt using three inputs: a PDDL domain file, legality constraints implemented as Python tests, and subset constraints specified as first-order logic (FOL) formulas.

\paragraph{Legality Constraints.}
Legality constraints only need to be defined once per-domain.
Hence, we implement them as Python tests that enable the efficient verification of legality. 
This also allows users to implement informative error messages and reusable code utilities that support the LLM's code generation.
\name requires the legality constraints to be provided as a standalone Python script containing a method \texttt{verifyLegality}, which inputs an instance file and returns a Boolean and an optional list of error messages.

\paragraph{Subset Constraints.}
Subset constraints are chosen according to the practical setting, and thus are changed frequently and may even be computed automatically.
Hence, we specify them declaratively using FOL formulas.
\name requires the subset constraints to be provided as a list of formulas in a PDDL-like syntax that supports standard FOL operators such as universal and existential quantification, XOR, and equality.
Additionally, we support the transitive closure and integer arithmetic.
Predicates must be suffixed with ``$I$'' or ``$G$'' if they refer to the initial state or the goal conditions, respectively. 
Predicates suffixed with ``$G$'' are new predicate symbols representing goal facts alongside the initial-state predicates. 
Their truth value is induced by the positive ground atoms listed in the goal conditions, interpreted under the closed-world assumption.
For example, in the Blocksworld domain, the subset constraint that in the initial state all blocks must be on the table and that in the goal all blocks must be in a single tower can be specified as
\begin{lstlisting}[style=pddlconstraint]
;; all blocks initially on the table
(forall (?b - block) (on-table_initial ?b))

;; only one clear block in the goal
(exists (?b1 - block)
        (and (clear_goal ?b1)
             (forall (?b2 - block)
                     (implies (clear_goal ?b2) 
                              (= ?b1 ?b2)
                                     )))).
\end{lstlisting}
\name also supports introducing new auxiliary predicates with the suffix ``new'' to define subset constraints more easily.
These predicates are syntactic sugar and the LLM is explicitly told that they must never occur in generated instances.
Further, \name supports defining and verifying subset constraints on the goal conditions under the open-world assumption, which is relevant for domains where goals are not conjunctions of ground positive atoms.

\paragraph{Prompting.}
Given the PDDL file, legality constraints, and subset constraints, we prompt the LLM to implement a standalone Python module that efficiently generates diverse instances that fulfill the legality and subset constraints.
This instance generator must consist of a single class with a predefined name and must expose a method \texttt{generateInstanceForSize}, which inputs an instance size, i.e, the number of non-constant objects, and an optional random seed, and returns either a string, corresponding to an instance file, or \texttt{None}, when no instance of the requested size exists. 
For example, on some domains, there is a minimum number of objects that each instance must contain. 
Lastly, we provide the LLM with a list of functional requirements that its generator must fulfill. 
These requirements cover what is being tested by the tests shown in Table~\ref{tab:generator-bugs}.
Example prompts can be seen in the Appendix~\ref{app:prompting}.

\subsection{Testing Instance Generators}
\label{sec:testing}

\begin{table*}[t!]
\centering
\small
\renewcommand{\arraystretch}{1.3}
\begin{tabular}{@{}
  >{\raggedright\arraybackslash}p{2.6cm}
  >{\raggedright\arraybackslash}p{3.2cm}
  >{\raggedright\arraybackslash}p{5.2cm}
  >{\raggedright\arraybackslash}p{3.8cm}@{}}
\textbf{Bug Category} & \textbf{Test} & \textbf{Description} & \textbf{Error Message} \\
\midrule
\multirow[t]{3}{=}{Code Bugs}
  & Class Loading Test
  & Try loading the generator class.
  & Python error message. \\
  
  & Instance Generation Test
  & Check whether instance generation terminates without a runtime error.
  & Python error message. \\
  
  & Efficiency Test
  & Check whether instance generation exceeds the time limit.
  & ``Generation exceeded the time limit of $N$ seconds.'' \\
  
\midrule
\multirow[t]{2}{=}{Instance File Bugs}
  & Parsing Test
  & Check whether the instance file can be parsed.
  & Error message of PDDL parser. \\
  
  & Instance Size Test
  & Check whether the number of objects in the instance matches the requested size.
  & ``Expected $N$ objects, but got $M$ instead.'' \\
  
\midrule
\multirow[t]{2}{=}{Quality Bugs}
  & Goal Fulfilled Test
  & Check whether the initial state already satisfies the goal.
  & ``The initial state already fulfills the goal.'' \\
  
  & Solvability Test\footnotemark[1]
  & Check whether the initial state's $h^{\text{FF}}$ value is infinite.
  & ``The initial state has the heuristic value $h^{\text{FF}}(s) = \infty$.'' \\
  
\midrule
\multirow[t]{2}{=}{Constraint Bugs}
  & Legality Test
  & Execute the given Python legality tests.
  & Error messages defined in the legality tests. \\
  
  & Subset Test
  & Verify whether the instance fulfills the given subset constraints using Z3.
  & Unsatisfiability core provided by Z3. \\
\end{tabular}
\caption{Overview of bug categories, tests, and their error messages for LLM-generated instance generators.
\footnotemark[1]An instance with a finite $h^{\text{FF}}$ value may still be unsolvable. 
}
\label{tab:generator-bugs}
\end{table*}

We define a sequence of tests for checking whether an LLM-generated instance generator executes correctly and generates sound instances.
These tests distinguish between four categories of bugs that are to be expected from LLM-generated instance generators: 
1. \emph{Code bugs} correspond to the generator throwing runtime errors or running too inefficiently.
2. \emph{Instance file bugs} correspond to a returned instance file being fundamentally incorrect. 
3. \emph{Quality bugs} correspond to a returned instance being trivial or impossible to solve.
4. \emph{Constraint bugs} correspond to a returned instance violating the legality or subset constraints.
Table~\ref{tab:generator-bugs} shows descriptions of all tests and the error messages they provide.

Given an instance generator, we attempt to generate $M$ instances for each instance size $n$ from a predefined set of sizes $N$.
For each generation attempt, we run the tests shown in Table~\ref{tab:generator-bugs} in order from top to bottom, skipping any remaining tests if a code or instance file bug occurs. 
If a test fails, we record the bug category, the inputs to the \texttt{generateInstanceForSize} method, the instance (if one was generated), and the error message.
Instances that pass all tests are guaranteed to be sound and are stored to later compute the diversity of the generator.
In practice, we run the testing process for multiple instance sizes in parallel.

\paragraph{Verifying Subset Constraints.}
To verify that a generated instance fulfills the subset constraints, which are given as FOL formulas, we use the theorem prover Z3~\cite{de2008z3}.
To do so, we translate the instance and the constraints into a finite Z3 model and let Z3 decide satisfiability.
Our translation encodes each object type as a custom \texttt{Enum} sort and each predicate as an uninterpreted Boolean function.
For every predicate, we instantiate all its ground atoms and add them to the model as axioms with truth values induced by the facts specified in the instance's initial state and goal conditions.
We then add the subset constraints to the model and let Z3 decide satisfiability.
If the model is unsatisfiable, the generated instance violates the subset constraints, corresponding to a constraint bug.
In this case, we store Z3's unsatisfiability core as an error message to provide feedback to the LLM.
This core is a subset of the axioms used in the unsatisfiability proof and gives the LLM concrete feedback about which facts cannot be satisfied together.

\subsection{Generator Improvement}
\label{sec:improvement}
Upon completion of the testing process, we generate reports about the LLM-generated instance generator's bugs and diversity.
These reports are then used to prompt the LLM again, asking it to generate an improved generator.

\paragraph{Bug Report.}
We inform the LLM about the mistakes in its instance generator without overloading the context window.
From all failed generation attempts, we select a small subset of generation attempts that covers all unique tests that have failed.
Using this set, we generate a bug report that, for each generation attempt, states the inputs to \texttt{generateInstanceForSize}, the incorrect instance file (if one was generated), and a list of the failed tests and their corresponding error messages. 

\paragraph{Diversity Report.}
We encourage that the instance generator returns diverse instances, as otherwise the LLM could minimize the number of bugs by implementing a generator that, for each instance size, only returns a single predefined sound instance.
Similar to the work of \citet{nunez2024nesig, nunez2025automated}, we define diversity as the distance between instances in a feature space.
For every generated sound instance, we compute a set of instance features: 1. the number of objects per type, 2. the number of atoms in the initial state per predicate, 3. the number of goal conditions per predicate, and 4. the $\hff$ heuristic value of the initial state~\cite{hoffmann2001ff}.
The diversity report then states the mean, median, standard deviation, and range for each feature.

\paragraph{Improvement Prompt.}
The improvement prompt begins by stating the number of generation attempts, how many of them had bugs, how many sound instances were generated, and how often the generator returned \texttt{None}. 
Next, we provide the bug report and ask the LLM to fix all bugs without introducing new ones.
Further, we provide the LLM with the diversity report and ask it to investigate whether the feature statistics have reasonable values or whether the diversity can be improved.
Lastly, we inform the LLM about the average runtime of its generator and ask it to consider whether it can be made more efficient.
However, if less than 50\% of the generated instances are sound, the improvement prompt skips the diversity and runtime reports since the LLM should solely focus on fixing its generator.
The prompt is then given to the LLM with the inclusion of all past prompts and responses.
The resulting new generator is then tested again as the improvement loop repeats.

\subsection{Final Generator Selection}
\label{sec:selection}
After the improvement loop exhausts a predefined time limit, we gather the generators that returned the largest number of sound instances and then select the generator that returned the most diverse instances.
This requires computing a diversity score for every considered generator.
Given the feature vectors of the sound instances of each generator, which were computed after testing, as described in Section~\ref{sec:improvement}, we first apply a shared pre-processing step, and then compute individual diversity scores.

\paragraph{Pre-processing.}
The pre-processing step begins by pooling all feature vectors into a single dataset and standardizing each dimension to ensure that the features are on the same scale. 
Afterward, we project the feature vectors to a lower-dimensional space using principal component analysis (PCA)~\cite{jolliffe2016principal}, retaining $95\%$ of the explained variance. 
This step removes redundant correlated dimensions, as otherwise the same structural difference could be counted multiple times when computing distances between feature vectors. 
For example, assume that we are given a Blocksworld instance and create a new one by stacking one previously tabled block onto another.
This replaces an \texttt{on-table} fact with an \texttt{on} fact and removes the \texttt{clear} fact of the supporting block. 
Thus, a single structural change affects the counts of three predicates (and potentially the $\hff$ value), causing the same difference to be represented across multiple correlated feature dimensions.

\paragraph{Diversity Scores.}
After the shared pre-processing step, we compute the diversity score of each generator individually.
For a generator $i$, we gather the corresponding instance feature vectors and group them by instance size $n \in N$, where $N$ is the set of test sizes.
Within each group, we measure diversity using the mean pairwise Euclidean distance ($\mpd$) between the projected feature vectors.
If fewer than two feature vectors are available for a size, we set $\mpd=0$.
We compute diversity only between instances of the same size because changing the instance size may change the instance features without introducing significant structural changes.
For example, a Blocksworld instance consisting of a single tower of 10 blocks has 9 \texttt{on} atoms, whereas an instance with a single tower of 20 blocks has 19 \texttt{on} atoms. 
The distance computed between these instances would therefore be large primarily because of their different sizes, not because they are structurally different. 
Computing the diversity per instance size removes this confounding effect.
Lastly, we compute the diversity score as the mean over the test sizes $ D_i=\frac{1}{|N|}\sum_{n\in N}\operatorname{MPD}_{i,n}.$

\paragraph{Deployment.}
When deploying the selected instance generator, every generation attempt must pass the tests defined in Table~\ref{tab:generator-bugs}, meaning that we only return an instance if the generator executed without errors and the instance is verified to be sound.
Otherwise, we return \texttt{None}.


\section{Experiments}
\label{sec:experiments}

We empirically show that 1. our \name approach can efficiently generate sound and diverse instance generators, and 2. the resulting generators can efficiently generate large numbers of instances of large sizes. 
We consider the domains from the International Planning Competition (IPC) 2023 learning track~\cite{taitler20242023}.
For these domains, we specify unique subset constraints that demonstrate \name{}'s ability to verify constraints of varying complexity.
Given the space limitations, we provide the description and subset constraint of each domain in the Appendix~\ref{app:domains}.
For the legality constraints, we implemented Python tests that check legality based on the formal specifications by \citet{grundke2025domain} and the implicit constraints encoded in the instance generators used for the IPC.
All experiments were run using an Apple M5 Max CPU with 128 GB of memory.

\subsection{Generating Instance Generators}
\label{sec:generating-generators}

\begin{table}[h]
\centering
\small
\begin{tabular}{lrrr}
Domain & Iter. & Sound & Diversity Relative to IPC \\
\toprule
Blocksworld & 3 & 100 & +836.7 \\
Blocksworld$^{S}$ & 1 & 100 & -100.0 \\
\midrule
Childsnack & 22 & 100 & +34.8 \\
Childsnack$^{S}$ & 14 & 100 & -44.7 \\
\midrule
Ferry & 29 & 100 & +41.8 \\
Ferry$^{S}$ & 11 & 100 & +11.5 \\
\midrule
Floortile & 8 & 100 & -42.3 \\
Floortile$^{S}$ & 16 & 100 & -52.6 \\
\midrule
Miconic & 7 & 100 & +48.7 \\
Miconic$^{S}$ & 7 & 100 & +38.9 \\
\midrule
Rovers & 14 & 100 & -25.1 \\
Rovers$^{S}$ & 5 & 100 & -2.8 \\
\midrule
Satellite & 23 & 100 & -9.3 \\
Satellite$^{S}$ & 12 & 100 & -15.2 \\
\midrule
Sokoban & 1 & 100 & +171.3 \\
Sokoban$^{S}$ & 18 & 100 & -84.9 \\
\midrule
Spanner & 5 & 100 & +10.8 \\
Spanner$^{S}$ & 19 & 100 & +3.7 \\
\midrule
Transport & 12 & 100 & -25.9 \\
Transport$^{S}$ & 13 & 100 & -62.4 \\
\end{tabular}
\caption{Soundness \& diversity of \name{}'s generators.
We show the increase or decrease (+\textbackslash-) in diversity w.r.t. the IPC generators.
$^S$ indicates the usage of subset constraints, which can enforce a lower diversity by restricting the instance subset.}
\label{tab:generator-synthesis}
\end{table}

In this experiment we show that \name can efficiently generate sound and diverse instance generators.
We generate two instance generators for each domain: One that only fulfills the domain's legality constraints and one that additionally fulfills subset constraints.
We use OpenAI's GPT-5.4-nano model with low reasoning and set a time limit of 10 minutes.
The testing process makes 20 generation attempts for each instance size $n \in \{15, 18, 22, 29, 38, 40, 44, 49, 54, 60\}$, totaling 200 test instances.
On the complex Sokoban domain, we use GPT-5.4-mini and test sizes $n \in \{38, 67, 83, 103, 123, 147, 171, 199, 227, 258\}$.
The total API cost was approximately 4\$.

To evaluate soundness, we compute for each generator the percentage of sound instances that were generated during testing.
To evaluate diversity, we compare each generators' diversity against the diversity of the generator of the same domain that was used in the IPC.
The diversity of the IPC generators was computed by generating the same number of instances for the test sizes used by \name, and then computing the diversity equivalently as described in Section~\ref{sec:selection}.
For each domain, Table~\ref{tab:generator-synthesis} shows the generators that were selected according to the criteria defined in Section~\ref{sec:selection}.
The superscript $^S$ indicates that the generator fulfills subset constraints.
For every generator, we show the iteration it was computed (``Iter.''), the percentage of generated sound instances (``Sound''), and the relative difference in diversity compared to the IPC generator (``Diversity Relative to IPC'').

Looking at the results, we see that within the short time limit of 10 minutes, \name computed generators that only returned sound instances for all domains.
Most generators were selected after at least 10 iterations of improvement because \name computed a generator with 100\% soundness after 1 to 3 iterations and then spent the remaining iterations improving diversity.
Considering the generators without subset constraints, we see that our generators are more diverse than the handcrafted IPC generators on 6 domains. 
For example, on Blocksworld, our generator produces a large variation in the number of towers, whereas the IPC generator is biased toward a small number of towers. 
On Ferry, our generator allocates cars to locations according to different patterns, whereas the IPC generator allocates cars uniformly.
We see the biggest decrease in diversity on the complex Floortile domain, where our generator creates only one robot and varies the grid dimensions less than the IPC generator.
Considering the generators with subset constraints ($^S$), we see that they are less diverse than the handcrafted IPC generators on 7 domains.
This is by design, as the generators must generate much smaller instance subsets than the IPC generators.
For example, on Blocksworld$^{S}$, we consider the instance subset where, in the initial state, all blocks must be on the table and, in the goal, all blocks must be in a single tower, enforcing a diversity score of $0$.
On Sokoban$^{S}$, we consider the subset where every box goal location is adjacent to the goal location of another box, drastically reducing the number of possible goal configurations.
On the other hand, on Miconic$^{S}$, the subset constraint only requires that every passenger's destination is above their origin floor, allowing our generator to sample the number of floors and allocate passengers with more variety than the IPC generator.
Importantly, the diversity scores are larger than $0$ on all domains, except Blocksworld$^{S}$, showing that \name{}'s diversity feedback prevents the LLM from computing generators that, for every instance size, return the same predefined sound instance.

\subsection{Generating Instances}

\begin{table}[h]
\centering
\small
\begin{tabular}{lrrrrrr}
\multicolumn{1}{c}{} & \multicolumn{2}{c}{Easy} & \multicolumn{2}{c}{Medium} & \multicolumn{2}{c}{Hard} \\
\cmidrule(lr){2-3} \cmidrule(lr){4-5} \cmidrule(lr){6-7}
Domain & \name & Gr25 & \name & Gr25 & \name & Gr25 \\
\toprule
Blocksworld & 100 & 93 & 100 & 0 & 100 & 0 \\
Childsnack & 100 & 97 & 100 & 3 & 97 & 0 \\
Ferry & 100 & 100 & 100 & 90 & 100 & 0 \\
Floortile & 100 & 10 & 100 & 0 & 100 & 0 \\
Miconic & 100 & 100 & 100 & 67 & 100 & 0 \\
Rovers & 100 & 100 & 100 & 10 & 97 & 0 \\
Satellite & 100 & 100 & 100 & 67 & 100 & 0 \\
Spanner & 100 & 100 & 100 & 0 & 100 & 0 \\
Transport & 100 & 100 & 100 & 87 & 100 & 0 \\
\bottomrule
Total time & \textbf{21s} & 7h & \textbf{2m} & 43h & \textbf{1h} & -- \\
\end{tabular}
\caption{
Percentages of successfully generated IPC instances per difficulty.
The bottom row shows the total runtime of successful generation attempts.
}
\label{tab:ipc-test-set-recreation}
\end{table}

\begin{table}[h]
\centering
\small
\begin{tabular}{lrrr}
Domain & Easy & Medium & Hard \\
\toprule
Blocksworld$^{S}$ & 100 & 100 & 100 \\
Childsnack$^{S}$ & 100 & 100 & 100 \\
Ferry$^{S}$ & 100 & 100 & 100 \\
Floortile$^{S}$ & 100 & 100 & 93 \\
Miconic$^{S}$ & 100 & 100 & 100 \\
Rovers$^{S}$ & 83 & 100 & 43 \\
Satellite$^{S}$ & 100 & 100 & 100 \\
Spanner$^{S}$ & 100 & 100 & 73 \\
Transport$^{S}$ & 100 & 100 & 100 \\
\bottomrule
Total time & 26s & 16m & 9h \\
\end{tabular}
\caption{
Percentages of successfully generated IPC instances per difficulty when using subset constraints.
The bottom row shows the total runtime of successful generation attempts.
}
\label{tab:ipc-test-set-recreation-constrained}
\end{table}

\begin{table}[h]
\centering
\small
\begin{tabular}{lrrr}
Domain & I2G & I2G$^{S}$ & NeSIG \\
\toprule
Blocksworld & \textbf{6s} & \textbf{6s} & 26s \\
Miconic & \textbf{6s} & \textbf{6s} & 26s \\
Satellite & \textbf{6s} & \textbf{6s} & 29s \\
Sokoban & \textbf{6s} & 7s & 221s \\
\end{tabular}
\caption{
Time needed to generate 100 instances from NeSIG's training sizes.
$^S$ indicates the usage of subset constraints.
}
\label{tab:nesig-comparison}
\end{table}

In this experiment, we show that \name's instance generators can efficiently generate large numbers of sound instances with large sizes.
To do so, we compare \name against the answer set programming approach by \citet{grundke2025domain}, which we refer to as ``Gr25'', and the RL approach by \citet{nunez2024nesig, nunez2025automated}, which we refer to as ``NeSIG'', using the publicly available data and results of both approaches.

\paragraph{Comparison to Gr25.}
\citet{grundke2025domain} evaluated the efficiency of their approach by ``recreating'' the IPC 2023 learning track test sets (excluding Sokoban), i.e., for every instance in the IPC test sets, they attempt to generate an instance of the same size with a 30 minute time limit per-instance.
For each domain, the IPC test sets are split into three difficulties, with each subset consisting of 30 instances.
Here, a higher difficulty means a larger instance size, with average sizes of $33$, $236$, and $1125$ for easy, medium, and hard instances, respectively.
For our comparison, we recreate the IPC tests using the instance generators from Section~\ref{sec:generating-generators}.
Table~\ref{tab:ipc-test-set-recreation} shows the percentages of successfully generated instances per difficulty, for Gr25 and our instance generators without subset constraints.
The bottom row shows the total runtime of successful generation attempts per difficulty.
We see that \name scales drastically better than Gr25.
For example, our generators recreated all medium difficulty IPC test sets within 2 minutes, whereas Gr25 required 43 hours to generate less than half of all medium instances.
Table~\ref{tab:ipc-test-set-recreation-constrained} shows the results for recreating the IPC test sets using our instance generators with subset constraints.
We see that, even with the additional verification of the subset constraints, our generators are still drastically more efficient than Gr25.

\paragraph{Comparison to NeSIG.}
\citet{nunez2024nesig, nunez2025automated} evaluated NeSIG's instance-generation policies by generating 100 instances with the same instance sizes as those used during their training.
For our comparison, we compute the maximum size of their generated instances for every shared domain and generate 100 instances using the generators from Section~\ref{sec:generating-generators}.
This corresponds to a size of 14 for Blocksworld, a size of 15 for Miconic, a size of 13 for Satellite, and a size of 38 for Sokoban.
Table~\ref{tab:nesig-comparison} shows the total runtime per domain for NeSIG and \name, where \name{}$^S$ indicates that the generators fulfill subset constraints.
Our instance generators are clearly more efficient than NeSIG's instance-generation policies, as we only require about 6 seconds of runtime to generate 100 instances on all domains, whereas NeSIG takes between 26 and 221 seconds.
Computing our generators is also more efficient than training NeSIG's policies, since \name only required 10 minutes to compute the generators, whereas NeSIG's policy training required between 2 and 6 days.
We note that an exact comparison between \name and NeSIG is not possible because \name was designed to efficiently generate diverse instances that fulfill given constraints, whereas NeSIG was designed to learn how to generate diverse and difficult instances.
Thus, our generators are much more efficient, but NeSIG's policies can generate difficult instances without additional input.

\section{Case Study: Evaluating Generalized Plans}
In this section, we conduct a case study of instance generation for the purpose of evaluating generalized planning methods.
The goal of generalized planning is to synthesize a generalized plan that can solve any instance from a given domain.
There is a wide variety of generalized planning methods, ranging from feature-based abstractions for computing policies~\cite{bonet2019learning}, to graph neural network policies~\cite{staahlberg2022learning}, and approaches that synthesize programmatic policies using LLMs~\cite{silver2024generalized}.
Using \name we can, for the first time, efficiently evaluate the generalization capabilities of such methods on specific instance subsets without needing excessive manual coding work or generator-learning.

We compare selected methods following the ideas of \citeauthor{gros2025per}'s scaling behavior evaluation~\shortcite{gros2025per}.
For each size $n \in \{10, 19, 31, 48, 57, 66, 73, 86, 95, 100\}$, we generate $50$ instances using the generators from Section~\ref{sec:generating-generators}, and run the generalized plans with a 1 minute time limit.
For each generalized plan, we then compute a statistical estimate of the average coverage $\hat{C}_{n}$ per-size $n$ and its confidence interval. 

\begin{figure}[h!]
    \centering
    \begin{subfigure}[t]{0.9\linewidth}
        \includegraphics[width=\linewidth]{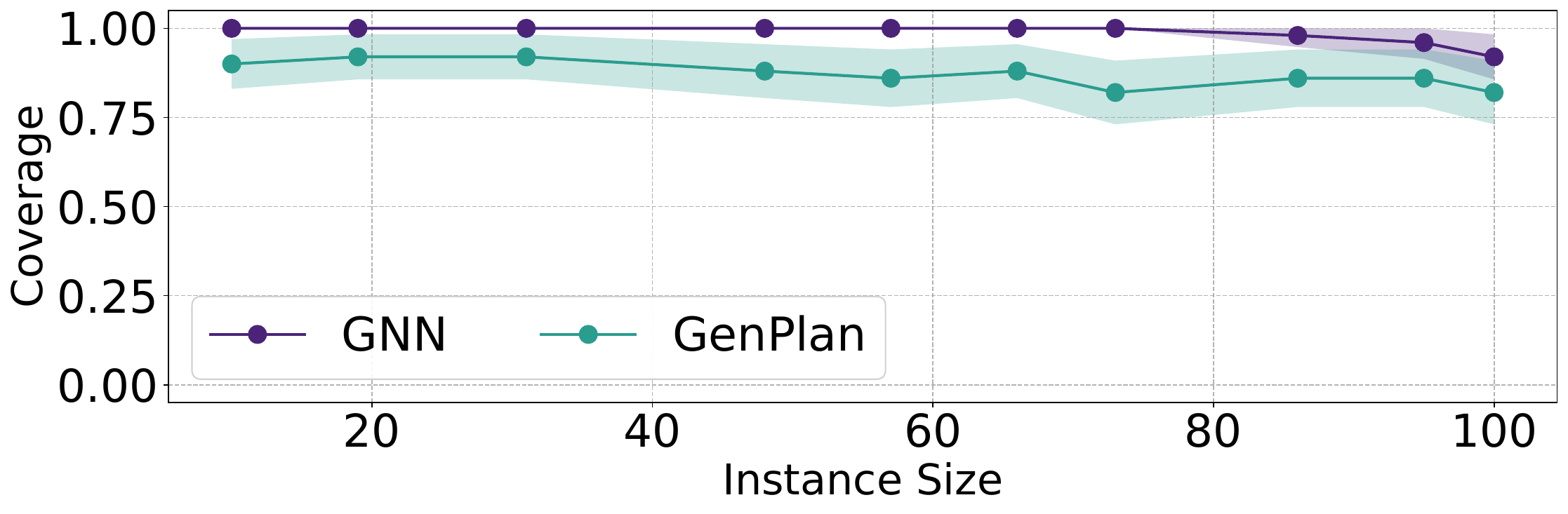}     
    \end{subfigure}
    
    \begin{subfigure}[t]{0.9\linewidth}
        \includegraphics[width=\linewidth]{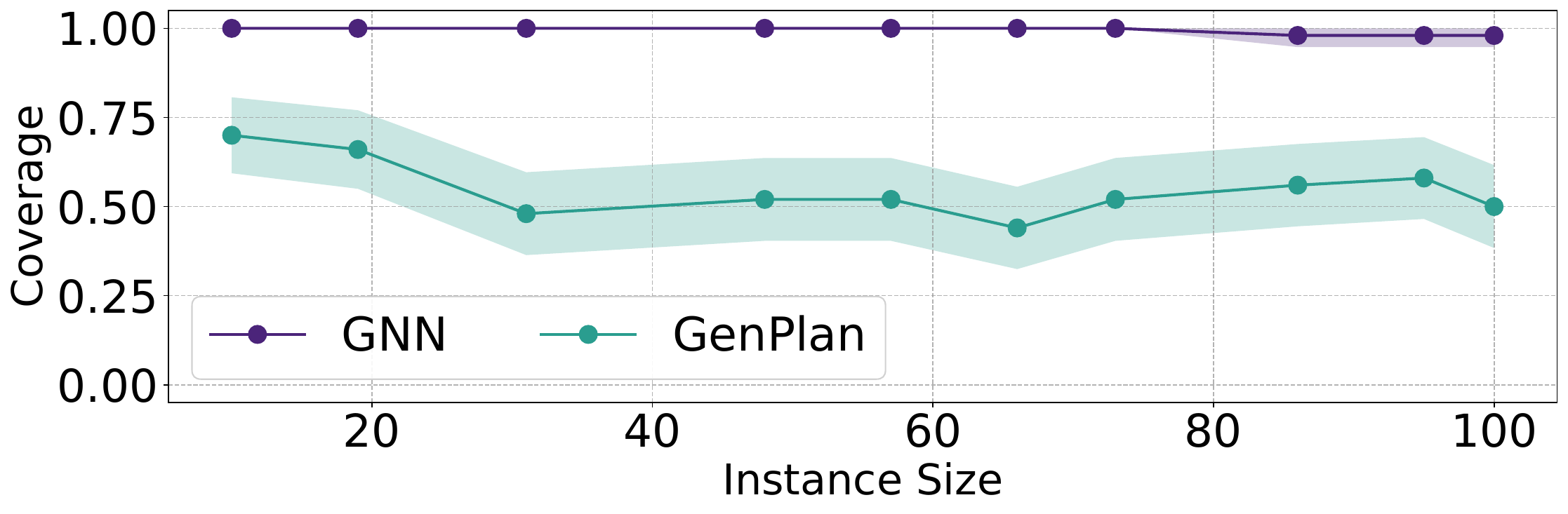}   
    \end{subfigure}
    
    \caption{Evaluation of GNN \& GenPlan policies on Satellite (\textbf{top}) and Satellite$^{S}$ (\textbf{bottom}).}
    \label{fig:satellite}
\end{figure}

\paragraph{Satellite}
We compare the GNN approach by \citet{muller2026qvalue}, which learns policies using graph neural networks, against the GenPlan approach by \citet{stein2026improved}, which uses LLMs to synthesize programmatic policies.
We evaluate these approaches on the Satellite generator without subset constraints and the Satellite$^{S}$ generator that fulfills the constraint that, in the initial state, all satellites have already allocated power to an instrument.
In Figure~\ref{fig:satellite}, we see that GNN and GenPlan achieve 80\% to 100\% coverage on the Satellite generator (top).
However, on the Satellite$^{S}$ generator (bottom), GenPlan's coverage drops significantly, whereas GNN's coverage remains at nearly 100\%.
This is because GenPlan's training data did not include instances from this instance subset, so its programmatic policy has a bug where it can fail if an instrument of any satellite is already powered on in the initial state.

\begin{figure}[h!]
    \centering
    \begin{subfigure}[t]{0.9\linewidth}
        \includegraphics[width=\linewidth]{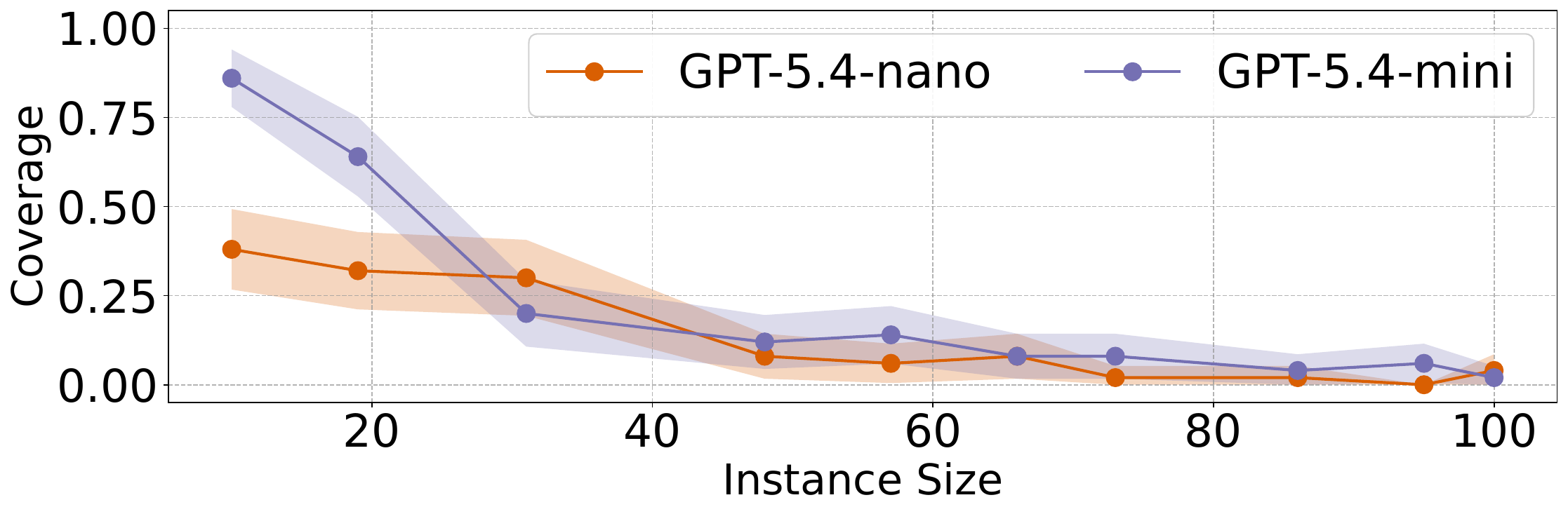}      
    \end{subfigure}
    
    \begin{subfigure}[t]{0.9\linewidth}
        \includegraphics[width=\linewidth]{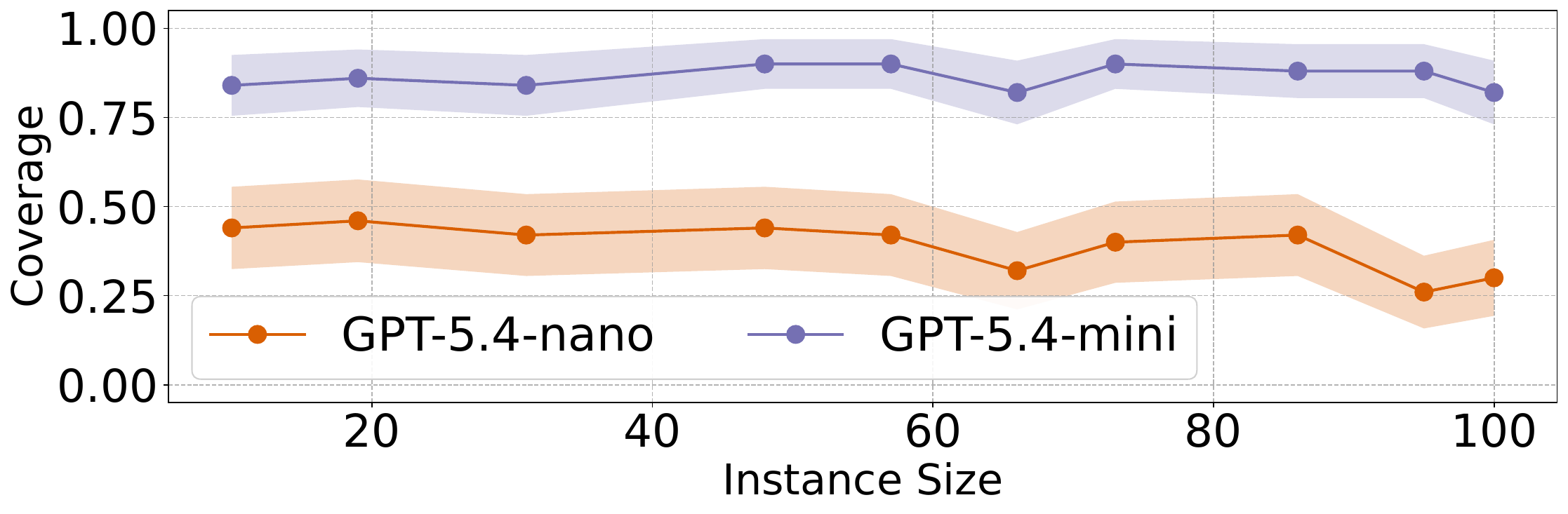}    
    \end{subfigure}
    
    \caption{Evaluation of GPT-5.4 models on Transport (\textbf{top}) and Transport$^{S}$ (\textbf{bottom}).}
    \label{fig:transport}
\end{figure}

\paragraph{Transport}
We evaluate OpenAI's GPT-5.4-nano \& mini models as generalized plans by prompting them to generate a plan for a given instance, where their reasoning effort is set to low.
We compare performance on the Transport generator without subset constraints against performance on the Transport$^{S}$ generator with subset constraints.
We consider the instance subset, where there is only one package that must be delivered which is to a location that is one or two roads away and there is a vehicle at the initial location of said package.
This means that the instances can be easily solved optimally with 3 or 4 actions and that most objects and facts are irrelevant to the goal.
In Figure~\ref{fig:transport} we see that on Transport (top) the coverage of both models quickly drops to near 0\% as the instance size increases.
However, on Transport$^{S}$ (bottom) GPT-5.4-nano's coverage is consistently below 50\%, whereas that of GPT-5.4-mini is above 75\%.
\name thus allows to easily exhibit, through varying the instance subset, the difference in planning-performance between these two language models.

\section{Conclusion}
\label{sec:conclusion}
Automating instance generation is a longstanding problem in the AI planning community.
In this work, we introduced \name, a highly efficient approach for generating instances that fulfill given constraints.
Given a PDDL domain, legality constraints implemented as Python tests, and subset constraints specified in FOL, a feedback loop between an LLM and a testing procedure iteratively constructs standalone Python generators that generate sound and diverse instances.
Our experiments show that \name can compute instance generators within minutes and that these generators scale substantially better than existing reasoning- and learning-based approaches.
Further, our case study on evaluating generalized planning methods showcases that \name enables effective and insightful experimentation with domain-instance subsets.
One interesting direction for future work is to use \name to automatically discover instance subsets where planning methods fail.
Given a set of instances where a planner failed, one could derive candidate subset constraints and then use \name to generate further instances for evaluation.

\bibliography{aaai2027}

\appendix
\section{Prompting}
\label{app:prompting}
In this section, we provide the prompts used for generating the initial instance generators and for improving the instance generators. We only show the templates of the prompts where all parts of the prompts that are specific to a particular domain or debugging iteration are omitted and replaced by `[...]'.

\subsection{Initial Prompt}
We use the following prompt for generating the initial instance generators:

\begin{lstlisting}[style=prompt]
You need to implement a standalone instance generator for the classical planning domain `[...]`.
The generator should take an instance size and generate legal instances of this domain.
Every generated instance must pass the provided legality test.
Optional subset constraints restrict the instances that the generator should return to a subset of all legal instances.

Here is the PDDL domain file of `[...]`:
```pddl
[...]
```

Constraint syntax:
In subset constraints, `_I` refers to the closed-world initial state, `_G` to explicitly specified goal facts under a closed-world interpretation, and `_G_open` to the goal under an open-world interpretation. Unsuffixed predicates refer to the initial state. `_new` predicates are auxiliary relations used only inside constraints. These suffixes never occur in PDDL instances.
In addition to first-order logic, constraints may contain these integer expressions: The expression '(count (?x - type) FORMULA)' evaluates to the number of objects of the given type that fulfill FORMULA. + is the sum of integer expressions. Integer expressions can be compared using =, <, <=, >, and >=.
The language also supports transitive closure for binary predicates: `(tc PREDICATE ?arg1 ?arg2)`.

Legality test:
Every generated instance must pass this Python legality test:

```python
[...]
```

Optional subset constraints:
Constraint 1: Context: [initial state/specified goal facts (closed world)], Formula: `[...]`
[...]

Your generator must fulfill these functional requirements:
- Legality: All generated instances must be legal instances of the `[...]` domain, as specified by the corresponding domain file. Only use types for objects if the domain contains the `:typing` requirement. All generated instances must pass the provided legality test.
- Subset constraints: If subset constraints are provided, all generated instances must satisfy them.
- Instance size: The generator should input the instance size, then randomly determine the number of non-constant objects per type such that the number of non-constant objects matches the given size. Domain constants do not count toward instance size. If no instance of the given size can be generated, the generator should return `None`.
- Solvable: All generated instances must be solvable, i.e., there must exist a valid plan that achieves the goal from the initial state.
- Non-trivial: All generated instances must be non-trivial, i.e., the goal should not already be fulfilled in the initial state.
- Runtime Errors: The generator should not raise runtime errors for any valid input.
- Efficiency: The code should be efficient and always terminate within [...] seconds. If an instance cannot be generated, return `None`.
- Diversity: The generator should cover as much of the distribution of legal instances fulfilling the given constraint(s) as possible. 

Your generator must fulfill these API requirements:
- Write a complete Python module that defines exactly one class named `[...]Generator`.
- Class API:
  - `__init__()`
  - `generate_instance_for_size(self, size: int, seed: int = None) -> str`
- Determinism: Use a local RNG which is seeded if a seed is provided.
- Return either a complete valid PDDL instance string or `None` if an instance cannot be generated for the given size.

Your generator must fulfill these style requirements:
- Return ONLY Python code. No markdown, no explanation.
- Use only Python standard library imports.
- No file I/O in the generated class.
- The code should be as simple as possible.
\end{lstlisting}

\subsection{Improvement Prompts}
For the generator improvement loop, we use the following prompt structure where `[FEEDBACK]' is a placeholder for the actual feedback provided to the LLM:
\begin{lstlisting}[style=prompt]
Debugging Pipeline Iteration [...]:
The instance generator that you implemented was tested by attempting to generate [...] instances for each of the following instance sizes:
[...]
Out of all generation attempts, [...] attempts had bugs, `None` was returned [...] times, and [...] correct instances were returned.

[FEEDBACK]

Given this feedback, please address the above points by updating your instance generator.
Ensure that your generator returns instances for every size for which instances that fulfill the given constraints exist, and not just for the tested sizes.
Keep the code simple.
\end{lstlisting}

We differentiate between two types of feedback: 1. Feedback about the types of bugs observed during the testing of the generator and 2. feedback about the runtime of the generator and the diversity of generated instances. 
If less than 50\% of the generated instances are sound, we provide only the feedback about observed bug types and leave the diversity feedback empty. 
If all generated instances are sound, there are no bugs and we therefore only provide information about the diversity of the instances and the runtime of the generator. 
If the soundness threshold of 50\% is reached but not all instances are sound, we combine both feedback types into the final feedback and let the LLM fix the bugs and improve diversity and efficiency at the same time.

This is the template for the bug feedback: 
\begin{lstlisting}[style=prompt]
Below are representative test logs that cover all observed bug types:

Begin Test Log
ID: [...]
Instance size: [...]
Instance file:
[...]
Bug 1:
  Bug type: [...]
  Error message: [...]
End Test Log
\end{lstlisting}

This is the template for the feedback about the diversity and runtime:
\begin{lstlisting}[style=prompt]
The generator achieved [...]% soundness, counting buggy and `None` results as failures.
To assess the diversity of the instances returned by your generator, a set of features was computed for every instance and then the mean, standard deviation, median, and range was computed for every feature across all instances.
With the help of these feature statistics, please assess whether your generator can be adjusted to generate more diverse instances, while improving or maintaining soundness.
It is important that the generator returns sound instances that have large and meaningful variations in their initial states and goals, as it would be expected from a useful instance generator.
Do not just blindly try to optimize some feature statistics, rather use the statistics to help you design a diverse generator. 
In particular, the instances should not optimize diversity while at the same time being extremely easy to solve.
It is also forbidden to optimize diversity by adding duplicate facts and they will not be considered in the feature statistics!

- hFF value of the initial state: Mean [...], Median [...], Standard Deviation [...], Range [[...], [...]]
- Number of goal conditions with predicate `[...]`: Mean [...], Median [...], Standard Deviation [...], Range [[...], [...]]
[...]
- Total number of goal conditions: Mean [...], Median [...], Standard Deviation [...], Range [[...], [...]]
- Number of initial-state atoms with predicate `[...]`: Mean [...], Median [...], Standard Deviation [...], Range [[...], [...]]
[...]
- Total number of initial-state atoms: Mean [...], Median [...], Standard Deviation [...], Range [[...], [...]]
- Number of [...] objects: Mean [...], Median [...], Standard Deviation [...], Range [[...], [...]]
[...]
- object_features instance_size: Mean [...], Median [...], Standard Deviation [...], Range [[...], [...]]
- Total number of objects: Mean [...], Median [...], Standard Deviation [...], Range [[...], [...]]

On average, your instance generator required [...]s to generate an instance.
Please consider whether the generator is efficient enough.
\end{lstlisting}

\section{Domains}
\label{app:domains}
For our experiments, we use the domains from the IPC 2023 learning track \cite{taitler20242023}. 
Here, we provide a brief description of each of the ten domains as well as the description and formalization of the subset constraints we considered in our experiments.

\paragraph{Blocksworld.} 
Blocks need to be moved on a table from an initial arrangement into a goal arrangement, using only one arm. Blocks can be stacked on top of each other to form towers. The arm is always empty in the initial state, i.e., no block is being held. 

We consider the instance subset where every block is initially on the table, and the goal requires exactly one tower. This fixes the optimal plan length to $2(n-1)$, where $n$  is the instance size. This makes it easy to evaluate the quality of computed plans. It also requires every block to appear in a plan, meaning no object can be ignored.

\begin{lstlisting}[style=pddlconstraint]
;; All blocks start clear.
(forall (?b1 - object) (clear_I ?b1))

;; There is at most one clear block. Legality guarantees that a tower top exists.
(exists (?b1 - object) 
  (and 
    (clear_G ?b1) 
    (forall (?b2 - object) 
      (implies 
        (clear_G ?b2) 
        (= ?b1 ?b2)))))
\end{lstlisting}

\paragraph{Childsnack.} 
Sandwiches must be prepared and delivered to children while respecting gluten allergies. Sandwiches are prepared by combining bread and content portions, both of which may contain gluten. Trays are used to transport sandwiches from the kitchen to one of the three tables children may sit at. There are exactly enough bread and content portions to feed all children.

We consider the instance subset where there is only a single tray and occupied tables have exactly one allergic and one non-allergic child. This requires preparing sandwiches in batches and routing the tray between multiple tables.

\begin{lstlisting}[style=pddlconstraint]
;; A single shared tray.
(= (count (?t - tray) (= ?t ?t)) 
   1)
   
;; Every occupied non-kitchen place has exactly two children:
;; one allergic and one non-allergic.
(forall (?p - place)
  (implies
    (and
      (not (= ?p kitchen))
      (exists (?c - child) (waiting ?c ?p)))
    (and
      (= (count (?c - child)
           (and 
             (waiting ?c ?p)
             (allergic_gluten ?c)))
         1)
      (= (count (?c - child)
           (and 
             (waiting ?c ?p)
             (not_allergic_gluten ?c)))
         1))))
\end{lstlisting}

\paragraph{Ferry.} 
A ferry transports cars between locations. Each car has a starting location and a goal location, where start and goal must be different from each other. The ferry starts empty and can only transport one car at a time. 

We consider the subset of instances with a hub-and-spoke structure: The ferry starts at a central hub, there is at least one non-hub (spoke) location, and every car’s trip either starts or ends at the hub.
This represents a structured and nontrivial routing task where the hub-spoke pattern must be exploited.

\begin{lstlisting}[style=pddlconstraint]
(exists (?hub - location)
  (and
    ;; The ferry begins at the hub.
    (at-ferry_I ?hub)

    ;; Require at least one spoke.
    (exists (?s1 - location) (not (= ?s1 ?hub)))

    ;; Every car travels between the hub and a spoke.
    (forall (?c - car ?origin - location ?destination - location)
      (implies
        (and
          (at_I ?c ?origin)
          (at_G ?c ?destination))
        (or
          (= ?origin ?hub)
          (= ?destination ?hub))))))
\end{lstlisting}

\paragraph{Floortile.} 
Robots move over a rectangular grid and paint tiles in specified colors. Robots may only occupy tiles that are not painted yet and not occupied by another robot. They can move to orthogonally adjacent tiles and paint tiles directly above and below their current location. All rows must be fully painted, except for the bottom row, which must not be painted at all.

We consider the instance subset where there is exactly one robot and one of the grid's dimensions is at least twice as large as the other. 
The corresponding instances feature narrow grids where only a single robot is available, and thus painting the required cells requires careful movement to avoid entering a dead-end.
To formalize this constraint compactly, we introduce auxiliary predicates with the suffix ``new''.

\begin{lstlisting}[style=pddlconstraint]
;; Exactly one robot.
(= (count (?r - robot) (= ?r ?r)) 
   1)

;; Column headers are precisely the tiles with no tile above them.
(forall (?t1 - tile)
  (=
    (column-header_new ?t1)
    (forall (?t2 - tile)
      (not (up ?t2 ?t1)))))

;; Row headers are precisely the tiles with no tile to their left.
(forall (?t1 - tile)
  (=
    (row-header_new ?t1)
    (forall (?t2 - tile)
      (not (left ?t2 ?t1)))))

;; One grid dimension is at least twice the other.
(or
  (>=
    (count (?rh - tile) (row-header_new ?rh))
    (+ (count (?ch - tile) (column-header_new ?ch))
       (count (?ch - tile) (column-header_new ?ch))))
  (>=
    (count (?ch - tile) (column-header_new ?ch))
    (+ (count (?rh - tile) (row-header_new ?rh))
       (count (?rh - tile) (row-header_new ?rh)))))
\end{lstlisting}

\paragraph{Miconic.} 
An elevator transports passengers from their origin floors to different destination floors.

We consider the instance subset where every passenger’s destination is above their origin floor. This creates monotone upward traffic: agents can exploit a shared direction of travel, but must still decide when to board and depart passengers along the route. This tests whether agents recognize and use the ordering structure rather than treating passenger requests as arbitrary elevator traffic. 

\begin{lstlisting}[style=pddlconstraint]
(forall (?p - passenger ?origin - floor ?destination - floor)
  (implies
    (and
      (origin ?p ?origin)
      (destin ?p ?destination))
    (above ?origin ?destination)))
\end{lstlisting}

\paragraph{Rovers.} 
Rovers navigate between waypoints, collect soil and rock samples, take images, and communicate the resulting data from a single lander. Only some rovers have the capabilities for soil/rock analysis or imaging. Soil and rock samples are located at waypoints and must be analysed by rovers with matching capabilities. Rovers have a limited capacity for transportation. Rovers may also have multi-mode cameras on board that must first be calibrated using objectives that are only visible from certain waypoints. 

We consider the instance subset where every rover has exactly one of the three capabilities (soil, rock, or image), all capabilities are represented, every available soil and rock sample must be communicated, and every objective has exactly one requested imaging mode. 
This requires carefully coordinating specialized rovers to achieve all goals.

\begin{lstlisting}[style=pddlconstraint]
;; Every rover has exactly one scientific capability.
(forall (?r - rover)
  (or
    (and
      (equipped_for_soil_analysis_I ?r)
      (not (equipped_for_rock_analysis_I ?r))
      (not (equipped_for_imaging_I ?r)))
    (and
      (not (equipped_for_soil_analysis_I ?r))
      (equipped_for_rock_analysis_I ?r)
      (not (equipped_for_imaging_I ?r)))
    (and
      (not (equipped_for_soil_analysis_I ?r))
      (not (equipped_for_rock_analysis_I ?r))
      (equipped_for_imaging_I ?r))))

;; All three specialist roles occur.
(exists (?r - rover)
  (equipped_for_soil_analysis_I ?r))

(exists (?r - rover)
  (equipped_for_rock_analysis_I ?r))

(exists (?r - rover)
  (equipped_for_imaging_I ?r))

;; Every available soil sample must be analyzed and communicated.
(forall (?w - waypoint)
  (=
    (communicated_soil_data_G ?w)
    (at_soil_sample_I ?w)))

;; Every available rock sample must be analyzed and communicated.
(forall (?w - waypoint)
  (=
    (communicated_rock_data_G ?w)
    (at_rock_sample_I ?w)))

;; Every objective must be imaged in exactly one mode.
(forall (?o - objective)
  (exists (?m1 - mode)
    (and
      (communicated_image_data_G ?o ?m1)
      (forall (?m2 - mode)
        (implies
          (communicated_image_data_G ?o ?m2)
          (= ?m1 ?m2))))))
\end{lstlisting}

\paragraph{Satellite.} 
Satellites rotate between directions and use onboard instruments to acquire images in requested modes. Satellites can only power up one of their instruments at a time.
Instruments must first be calibrated by pointing in specified directions. The goal is to take at least one image in a specific mode and, optionally, let satellites point into specific directions. 

We consider the instance subset where all satellites have already allocated power to an instrument in the initial state.

\begin{lstlisting}[style=pddlconstraint]
(forall (?s - satellite)
  (not (power_avail ?s)))
\end{lstlisting}

\paragraph{Sokoban.} 
A robot moves through a grid and pushes boxes onto designated locations. The grid may contain impassable walls. Neither robots nor boxes may enter wall tiles. To push a box in a direction the robot must stand on the opposite side of a box.

We consider the instance subset where every box goal location is adjacent to the goal location of another box. 
This results in crowded goal locations for boxes, which makes moving the boxes difficult because the robot cannot enter cells occupied by boxes.

\begin{lstlisting}[style=pddlconstraint]
(forall (?b1 - box ?g1 - location)
  (implies
    (at_G ?b1 ?g1)
    (exists (?b2 - box ?g2 - location ?d - direction)
      (and
        (not (= ?b1 ?b2))
        (at_G ?b2 ?g2)
        (adjacent_I ?g1 ?g2 ?d)))))
\end{lstlisting}

\paragraph{Spanner.} 
A worker traverses a linear chain of locations, collects spanners, and uses them to tighten nuts. Each spanner is single-use. All nuts are loose in the initial state and must be tightened in the goal. The worker always starts at the first location, the shed, and all nuts are located at the last location, the gate. 

We consider the instance subset where there are equal numbers of spanners and nuts and there is at most one spanner at each location. This creates a limited-resource scenario requiring the worker to pick up and use every encountered spanner.

\begin{lstlisting}[style=pddlconstraint]
;; Every location strictly between shed and gate has at most one spanner.
(forall (?l - location)
  (exists (?s1 - spanner)
    (implies
      (at ?s1 ?l)
      (forall (?s2 - spanner)
        (implies
          (at ?s2 ?l)
          (= ?s1 ?s2))))))

;; Different spanners occupy different locations.
(forall (?s1 - spanner ?s2 - spanner ?l - location)
  (implies
    (and (at ?s1 ?l) (at ?s2 ?l))
    (= ?s1 ?s2)))

;; Every available spanner is necessary: one per nut.
(=
  (count (?s - spanner) (= ?s ?s))
  (count (?n - nut) (= ?n ?n)))
\end{lstlisting}

\paragraph{Transport.} 
Vehicles move packages through a road network from their initial locations to their destinations. Vehicles have variable capacity and start empty. The road graph is non-reflexive, connected and symmetric.

We consider the instance subset where only one package exists, with this package being initially colocated with a vehicle and its destination reachable by traversing at most two roads. 
This means the optimal plan length is always 3 or 4, and with an increasing instance size, an increasing number of irrelevant objects are added, as only the package's origin and destination, the colocated vehicle and its capacity, and a possibly intermediate location are relevant.

\begin{lstlisting}[style=pddlconstraint]
(exists
  (?lg - location ?pg - package ?li - location ?lm - location ?vi - vehicle)
    (and
      (at_I ?pg ?li)
      (at_I ?vi ?li)
      (or
        (and
          (not (= ?li ?lm))
          (not (= ?lm ?lg))
          (not (= ?lg ?li))
          (road_I ?li ?lm)
          (road_I ?lm ?lg))
        (and
          (not (= ?lg ?li))
          (road_I ?li ?lg)))
      (at_G ?pg ?lg)
      (forall (?lgp - location ?pgp - package)
        (implies
          (at_G ?pgp ?lgp)
          (and
            (= ?pg ?pgp)
            (= ?lg ?lgp))))))
\end{lstlisting}

\end{document}